%% file: neurips_2024.tex
\documentclass{article}

\usepackage[preprint]{neurips_2024}

\usepackage[utf8]{inputenc} %
\usepackage[T1]{fontenc}    %
\usepackage{hyperref}       %
\usepackage{url}            %
\usepackage{booktabs}       %
\usepackage{amsfonts}       %
\usepackage{nicefrac}       %
\usepackage{microtype}      %
\usepackage{xcolor}         %

\usepackage{epigraph}
\usepackage{xspace}
\usepackage{graphicx}
\usepackage{adjustbox}
\usepackage{enumitem}
\usepackage{wrapfig}
\usepackage{caption}
\usepackage{subcaption}
\usepackage{float}
\usepackage{tabularx}
\usepackage{xspace}
\usepackage{soul}
\usepackage{xcolor}
\usepackage{color, colortbl}
\usepackage{graphicx}
\usepackage{blindtext}
\usepackage{pythonhighlight}
\usepackage{algorithm}
\usepackage{algorithmic}
\usepackage{amsmath}

\definecolor{kellygreen}{rgb}{0.3, 0.73, 0.09}
\definecolor{alizarin}{rgb}{0.82, 0.1, 0.26}
\definecolor{myhighlight}{rgb}{0.804,0.98,0.77}
\usepackage{pifont}%

\newcommand{\method}{\textsc{ToolDec}\xspace}

\title{Don't Fine-Tune, Decode: \\Syntax Error-Free Tool Use via Constrained Decoding}

\author{%
  Kexun Zhang \thanks{Equal contribution. Correspondence to \texttt{kexun@cmu.edu}.} \\
  Carnegie Mellon University\\
  \texttt{kexun@cmu.edu} \\
  \And
  Hongqiao Chen $^{*}$ \\
  California Institute of Technology\\
  \texttt{harrychen@caltech.edu} \\
  \AND
  Lei Li \\
  Carnegie Mellon University\\
  \texttt{leili@cs.cmu.edu} \\
  \And
  William Yang Wang \\
  UC Santa Barbara\\
  \texttt{william@cs.ucsb.edu} \\
}

\begin{document}

\maketitle
\vspace{-15px}
\begin{abstract}
\input{sections/010_abs}
\end{abstract}

\vspace{-5px}

\begin{figure}[h!]
      \centering
	   \begin{subfigure}{0.9\linewidth}
		\includegraphics[width=\linewidth]{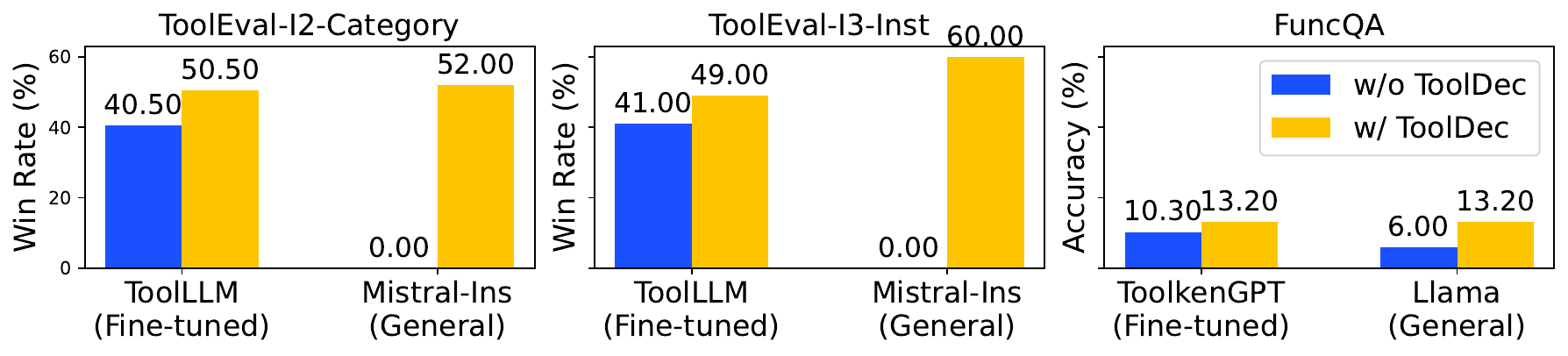}
	\end{subfigure}
    \vfill
	   \begin{subfigure}{0.9\linewidth}
		\includegraphics[width=\linewidth]{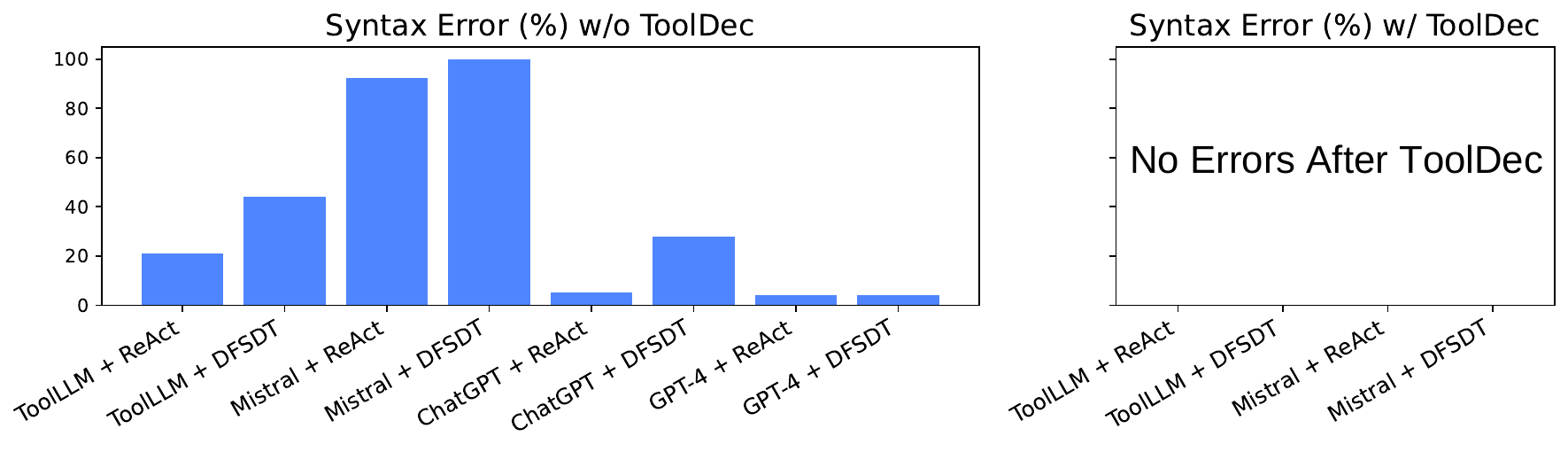}
		\end{subfigure}
 \caption{On various benchmarks, \method improves both fine-tuned specialist models (ToolLLM) and generalist models (Mistral-Instruct and Vicuna). Mistral-Instruct is improved from an initial 0\% to be even better than ToolLLM. \method also eliminates all syntax errors.}
\label{fig:fig1}
\end{figure}

\section{Introduction}

\input{sections/020_intro}

\section{Related Work}

\input{sections/021_related}

\section{Proposed Method: \method}
\input{sections/040_method}

\section{Experiment}
\input{sections/045_exp}

\section{Conclusion}
\input{sections/070_conclusion}

\bibliography{iclr2024_conference}
\bibliographystyle{icml2024}

\newpage
\appendix
\input{sections/100_app}

\end{document}

%% file: sections/010_abs.tex
Instruction-tuned large language models (LLMs) excel at many tasks but often fail to use external tools due to complicated and unfamiliar syntax constraints.
While extensive fine-tuning and prompting can mitigate the issue, these approaches are expensive and hard to generalize.
Furthermore, because syntax constraints are only learned implicitly during fine-tuning, models still make frequent syntax errors.
Motivated by the fact that these constraints can be better satisfied \textit{explicitly} with constrained decoding, we propose \method, a decoding algorithm using finite state machines to force LLMs to follow tool syntax.
Our experiments show that \method eliminates all syntax errors, achieving significantly better performance on various base models and benchmarks.
More surprisingly, when applied to generalist out-of-the-box LLMs such as Mistral-Instruct, \method improves its accuracy in tool use from the initial 0\% to an impressive 52\%, matching the performance of specialized fine-tuned models such as ToolLLM.
We release our code at \url{https://github.com/chenhongqiao/tooldec}.

%% file: sections/020_intro.tex
Augmenting large language models (LLMs) with external tools \citep{mialon2023augmented} enables them to solve complex problems. Current LLMs can utilize retrievers \citep{shen2023hugginggpt, Gupta2022VisProg, schick2023toolformer}, RESTful APIs \citep{qin2023toolllm, song2023restgpt}, program interpreters \citep{chen2022program, gao2023pal}, and various other tools.
As existing tools are being modified and new tools are being created every day, it is important for LLMs to be able to use unknown tools that are not in the training set.

\begin{figure}[h!]
  \centering
  \includegraphics[width=1\textwidth]{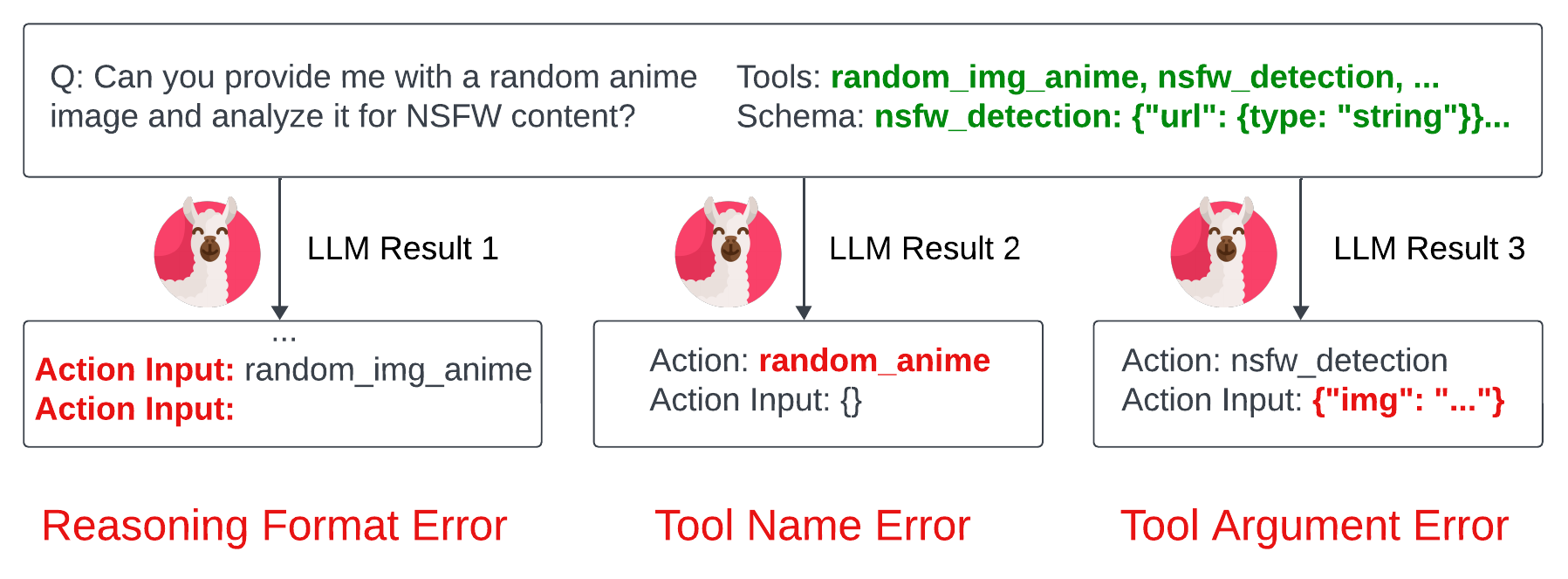}
  \vspace{-15pt}
  \caption{Common syntactical modes of failure of tool-use LLMs include reasoning format error, tool name error, and tool argument error. Even fine-tuned models have a significant level of syntax error.}\label{fig:failures}
\end{figure}

Out-of-the-box LLMs such as Mistral and Llama, even when instruction-tuned to be very capable on many other tasks \citep{jiang2023mistral}, can fail in using tools, as demonstrated by its low (and even \textit{zero}) performance in \autoref{fig:fig1}.
One major reason for their bad performance is syntax errors.
For example, Mistral-Instruct-7B has a syntax error rate of over 90\% when using some unknown tools in ToolEval \citep{qin2023toolllm}, which results in its 0\% accuracy.
Even very capable models such as GPT-4 make syntax errors on new tools.
Previous approaches use extensive fine-tuning or prompting \citep{qin2023toolllm, hao2023toolkengpt} to teach LLMs tool syntax, which reduces syntax errors but not all of them.
In \autoref{fig:fig1}, although the fine-tuned ToolLLM makes significantly fewer syntax errors than the general Mistral-Instruct, it still has an over 20\% error rate. We show examples of common modes of failure in \autoref{fig:failures}.

We argue that fine-tuning or prompting is neither optimal nor enough to enforce syntax constraints, because general instruction-tuned models already have a ``rough idea'' of what tool to use in different scenarios.
Fine-tuning or prompting approaches expect the model to learn and follow syntax constraints from tool use examples in training data or in-context demonstration.
However, these constraints can be explicitly modeled with symbolic rules.
Directly applying these rules in model generation can be more accurate than either fine-tuning or prompting, as they do not need extra compute or prompting while guaranteeing the model is syntax error-free.

To this end, we propose \method,
a decoding algorithm guided by a finite-state machine (FSM) to ensure LLMs invoke tools properly.
We automatically convert tool syntax schemas to an equivalent finite-state machine.
During decoding, \method transitions from state to state as decoding progresses.
At each decoding step, \method %
samples from the valid subset of tokens allowed by the tool syntax.%
This way, \method is able to always generate syntactically correct tool calls.
Since syntax constraints are enforced by the FSM, there's no need for them to appear in prompts.
We further use an LLM to simplify the tool description prompt by removing the syntax constraints from it.
\method is model-agnostic and can be combined with any base LLM.
More examples comparing \method and other tool LLMs can be found in Appendix \ref{app:example}.

We evaluate \method by applying it to 5 base models and evaluating it on 4 benchmarks.
On base models that are specifically fine-tuned for tool use purposes, such as ToolLLM \citep{qin2023toolllm} and ToolkenGPT \citep{hao2023toolkengpt},
\method significantly improves their accuracy by at most 21 points and greatly reduce the number of tokens in the output.
On base models that initially have zero accuracy on ToolBench \citep{qin2023toolllm} such as Mistral \citep{jiang2023mistral} and Vicuna \citep{vicuna2023}, \method improves their accuracy from 0 to at most 60\%.

Our contributions can be summarized as follows:

\begin{itemize}[leftmargin=10pt, nolistsep, noitemsep]
    \item We propose \method, a constrained decoding algorithm for LLMs to use tools without syntax errors.
    \method can off-load syntax constraints to an FSM, and eliminate all syntax errors.%
    \item We verify \method's superior performance by combining it with 5 base models and evaluating on 4 benchmarks.
    Our experiments show \method improves all base models significantly.
    \item We further compare \method on generalist models with specialized tool-using models. We find that \method, as an alternative, can achieve comparable performance while retaining the model's performance on other reasoning benchmarks.%
\end{itemize}

%% file: sections/021_related.tex
\textbf{Fine-tuning language models to use tools.}
Language models can be fine-tuned to use tools with data that contain interleaving text and tool use.
Earlier studies make language models use a single tool like a retrieval module \citep{borgeaud2022improving, guu2020retrieval} or a search engine \citep{nakano2021webgpt} by fine-tuning.
Recent advances in tool-augmented language models that use multiple tools \citep{schick2023toolformer, parisi2022talm} also fine-tune language models to use tools including QA models, translation models, calculators, and search engines.
ToolkenGPT \citep{hao2023toolkengpt} proposes to use several special tokens to represent tools and only tunes the embeddings of the tokens so that new tool adoption can be more efficient.
However, fine-tuning approaches cannot adapt to new tools without training data.

\textbf{In-context learning for tool use.}
Language models can learn from in-context examples \citep{brown2020language} and follow instructions \citep{ouyang2022training}.
This makes it possible to simply put the descriptions of tools in the prompt and ask language models to use them.
Recent works put tool documentation and demonstration in the prompt to use neural models \citep{shen2023hugginggpt}, RESTful APIs \citep{qin2023toolllm, song2023restgpt}, program interpreters \citep{chen2022program, gao2023pal} and many other tools to solve problems.
In-context learning does not need extra model tuning to use new tools.
However, the syntax and semantic constraints of new tools are entangled in the prompts, resulting in longer prompts and syntax errors.

\textbf{Constrained decoding and finite-state machines.}
Previous constrained decoding methods reduce the large search space of lexically constrained decoding with finite-state machines \citep{anderson-etal-2017-guided}, grouping together similar candidates \citep{hokamp-liu-2017-lexically}, and better search algorithms \citep{miao2019cgmh, lu2021neurologic, lu2022neurologic}.
However, lexical constraints are not expressive enough to regulate tool calls.
While finite-state machines have to be weighted and probabilistic to deal with the soft constraints in natural language \citep{eisner2002parameter, rastogi2016weighting}, the constraints for syntactic tool calls are hard constraints that are much easier for FSMs.
Grammar-constrained decoding has been used for structural NLP tasks such as code generation \citep{yin2017syntactic}, semantic parsing \citep{stengeleskin2024zero}, coreference resolution, POS tagging and many others \citep{geng2023grammar}.
\citet{willard2023efficient} uses finite state machines to guide LLMs to generate outputs efficiently and conform to grammar.

%% file: sections/040_method.tex
To use a tool, an LLM must first refer to an existent tool by its designated name. 
Then, it needs to generate arguments that adhere to the grammar of that tool (e.g. a JSON Schema). 
Motivated by the fact that it is easy to verify the syntax of a tool call using a finite-state machine (FSM), we propose \method, a LLM tool-use framework that uses an FSM to eliminate syntax errors.%
During each decoding step, the model samples from a subset of the vocabulary that only contains syntactically correct tokens. 
This subset is dictated by the current state of the FSM and the sampled token determines the next state to transition to.  

We designed an algorithm to automatically construct the described FSM from tool documentation recursively (Section \ref{sec:construction}). 
It takes advantage of machine-readable endpoint documentation (e.g., OpenAPI), which is very common in software engineering. 
In addition to constructing an FSM, we use another LLM to remove syntax constraints from the tool documentation (Section \ref{sec:compress}).
The goal of this step is to shorten the extensive prompts and demonstrate that FSM on its own is sufficient to guarantee syntactically correct tool calls.
An example of constrained decoding with \method is provided in Section \ref{sec:step_ex}.

\begin{figure}[h!]
  \centering
  \includegraphics[width=1\textwidth]{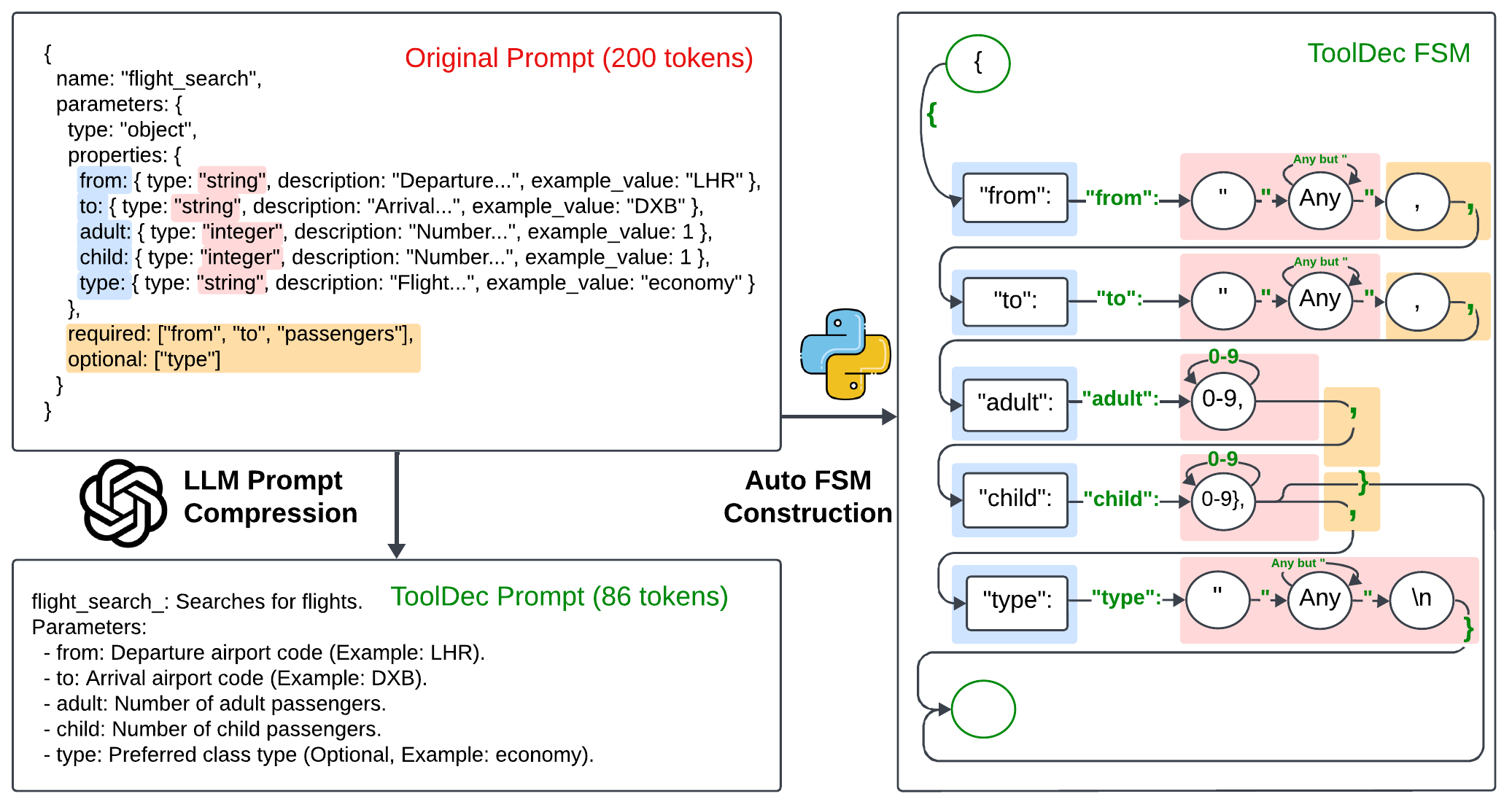}
  \caption{Converting a tool documentation to a simplified prompt and an FSM. The caption on each state in the FSM denotes the valid token set at that step. FSM will transition to the corresponding state when the token on that edge is generated by the LLM.
  }\label{fig:pipe}
\end{figure}

\subsection{Construction of \method FSM}
\label{sec:construction}
\textbf{Definition of \method FSM.}
An FSM is a 5-tuple $(S,V,g,s_0,R)$, consisting of a finite state set $S$, an alphabet $V$, a transition function $g: S\times V \rightarrow S$, an initial state $s_0$ and a set of accepting states $R$.
In our case, $S$ and $g$ are constructed from the tool documentation.
$V$ is the token vocabulary of the language model.
$R$ corresponds to pre-defined tokens that can determine the LM has completed the task, like `\textless{}EOS\textgreater{}'.

\textbf{FSM-constrained decoding.} At each decoding step $t$, \method maintains a current state $s$.
The LLM can only sample from the tokens permitted by the FSM, i.e. the tokens for which $g(s, \cdot)$ is defined.
These permitted tokens are a subset of $V$ and we denote them as $V_s$.
After generating one token $a$, \method transits to another state $g(s,a)$ specified by the FSM transition function.
The permitted tokens for each state can be the full vocabulary, or a valid subset corresponding to tool names and argument types.
With the next token, we move on to the next decoding step and transition the current state $s$ to the next state $g(s,a)$.
The pseudo-code of this algorithm is listed in Algorithm \ref{algo}.

\textbf{Constructing FSM from Tool Documentation.}
In this paper, we focus on the automatic construction of guidance FSM for REST APIs and Python functions.
With the following algorithm, we were able to cover 16,000+ tools across 4 different domains, representing various tool-use applications explored in previous studies \citep{qin2023toolllm, hao2023toolkengpt, yuan_easytool_2024}. 
This approach can be extended should future applications emerge because a finite state machine can be algorithmically constructed for any regular grammar.
Our FSM construction algorithm assumes that the syntax grammar of a tool is documented in a machine-readable format, for example, in the OpenAPI specification.

In \autoref{fig:pipe}, we illustrate an automatically constructed FSM from a JSON Schema.
The nodes represent the states.
The caption inside a state $s$ illustrates the valid tokens $V_s$ and the edges represent possible transitions based on the token sampled.
The generation starts at an initial state and stops with an accepting state (the green nodes).
For each parameter of a tool, we create a sub-machine for its syntax.
For example, the first row with 4 nodes in \autoref{fig:pipe} is the sub-machine for the parameter ``from''.
Each sub-machine has two parts -- one for parameter names and one for parameter values.
Name sub-machines (with blue backgrounds) only accept one string -- the name for a particular parameter.
Value sub-machines (with pink backgrounds) accept strings that follow the format of the parameter value.
For example, for the parameter ``adult'', it only accepts digit strings that end with a comma.
After generating the name and value of a parameter, an LLM might generate some end-of-generation tokens (with orange backgrounds, such as ` , ' or ` \} ') to move on to the next parameter.

For required parameters, their sub-machines must be passed through, while for optional parameters, there's a skip connection that allows the model to go around it without generating an optional parameter (for example, the parameter ``type'' in \autoref{fig:pipe}).
Note that this algorithm can also work for nested parameters (parameters that have subfields), because we can just apply the process recursively.%

Based on $V_s$, we pre-compute the token mask for every $s \in S$ during construction to allow $\mathcal{O}(1)$ filtering during inference. 
This construction process only needs to be executed once for every tool, and can be cached. 
The size of the FSM scales linearly as the number of arguments of a tool increases, which is the same as the number of tokens if prompting were used. 
However, our approach is much more efficient because its computation and memory overhead is negligible (less than 0.1\%) when compared to the GPU cost of LLMs.

\begin{figure}[h]
  \centering
  \includegraphics[width=1\textwidth]{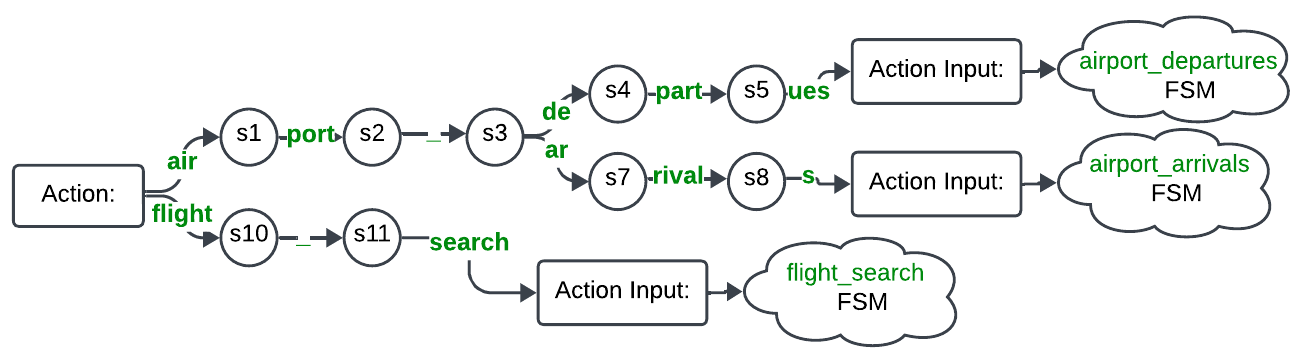}
  \caption{Linking multiple FSMs to guide the LLM through reasoning and tool selection. After a tool is selected, the FSM then transitions to start generating arguments.
  }\label{fig:trie}
\end{figure}

\textbf{Linking FSMs for Guided Reasoning and Tool Selection.}
Tool enhanced problem solving involves multiple steps of reasoning and selecting tools for each step. 
This can be guided by linking multiple FSMs together. 
For guided tool selection, we tokenized the names of all available tools and constructed a tree structure similar to a trie \citep{fredkin1960trie}. 
Every leaf node has a $g$ defined to transition to the $s_0$ of the FSM for that tool. 
An example can be seen in \autoref{fig:trie}. 
As the number of tools increases (for example, in the KAMEL \citep{kalo2022kamel} benchmark, toolset size reaches 234), the trie could automatically expanded by adding more nodes to the tree. 
Reasoning syntax, such as ReAct \citep{yao2023react}, can also be incorporated into the FSM as showns in \autoref{fig:trie}.

\subsection{Prompt Compression}
\label{sec:compress}
Since syntax constraints can be effectively handled by our decoding algorithm, we created a prompt compression pipeline to remove these constraints from the tool documentation. 
For each tool, we prompted an LLM to rewrite the tool documentation while ignoring the hierarchical structure of JSON schema and the typing of each parameter. 
Resulted was a simple list of parameters with concise descriptions. 
This list of parameters is necessary for the tool-use LLM to match the context information it has at hand with a tool. 
We show an example of the compressed prompt in \autoref{fig:pipe}. The prompt we used to rewrite the tool documentation can be found in Appendix \ref{app:prompt}.

By removing syntax information from the prompt, we reduce the number of tokens and show that syntax constraints are more efficiently handled by decoding algorithms rather than prompting. 

\subsection{Inferencing with FSM and Compressed Prompt}
\label{sec:step_ex}
Inferencing with \method involves using the \method FSM in conjunction with the compressed prompt. First, the compressed prompt, succinct and free of syntax constraints, is provided to the LLM to begin generation. At each step $t$, we do not directly sample from the next token distribution $P(x_t|x_{1..t-1})$ calculated by the LLM.
Instead, we zero out the probabilities of invalid tokens for which the transition function is undefined, and normalize the probabilities,
\begin{equation*}\small\tilde P(x_t=a)= \left\{
    \begin{aligned}&\tfrac{P(x_t=a|x_{1..t-1})}{\sum_{a'\in V_s}P(x_t=a'|x_{1..t-1})},& \exists g(s,a),\\ &0, & \text{otherwise}\end{aligned}\right..
\end{equation*}
The next token $a$ is then sampled from the modified distribution $\tilde{P}(x_t|x_{1..t-1},s)$.

\begin{figure}[h]
  \vspace{-5pt}
  \centering
  \includegraphics[width=1\textwidth]{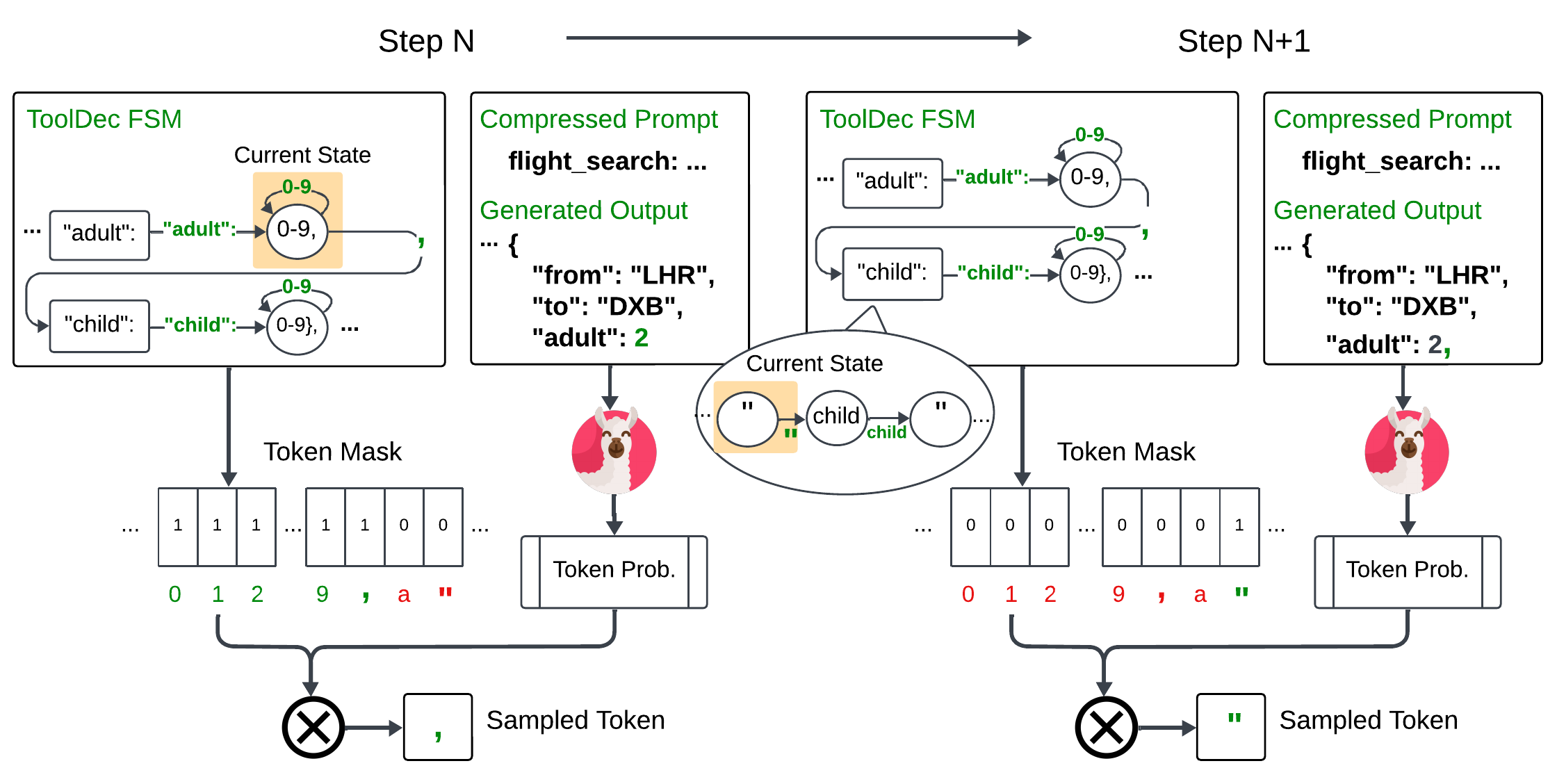}
  \caption{A decoding step using \method FSM. The invalid tokens at the current FSM state are masked out from the token probabilities.}\label{fig:inference}
  \vspace{-10pt}
\end{figure}

We show an example in \autoref{fig:inference}. First, the semantic description of flight\_search is provided to the LLM. At step $n$, the current state $s_n$ only permits the generation of integers or comma, which corresponds to adult's data type integer. By multiplying the token mask with $P$, we obtain $\tilde{P}$ where the probability of all other tokens are zeroed out. From the shifted probability, we sampled ` , '.
As we moves on to step $n+1$, the FSM also transitions to $s_{n+1}$ following $g(s_n, \text{` , '})$. Here with the zoom in bubble, we show the actual FSM that accepts the parameter name ``child'', which only accepts ` " ' as the first token. A similar process repeats until the LLM finishes generating a tool call. 

%% file: sections/045_exp.tex
As a model-agnostic decoding algorithm, \method can be applied to any LLM with token logits access.
We evaluate \method by applying it to 5 base LLMs on 4 benchmarks.
In this section, we first introduce the base LLMs (Section \ref{sec:ex_base}) and benchmarks we evaluate \method on (Section \ref{sec:bench}).
Then we report our main results (Section \ref{sec:main}) and further ablate why \method can be a good complement/alternative for fine-tuning and prompting (Section 
\ref{sec:abl}).

\subsection{Base LLMs}
\label{sec:ex_base}

We evaluate how \method can improve 5 LLMs --- ToolLLM \citep{qin2023toolllm}, ToolkenGPT \citep{hao2023toolkengpt}, RestGPT \citep{song2023restgpt}, Vicuna-7B \citep{vicuna2023}, and Mistral-7B-Instruct \citep{jiang2023mistral}.
The first three are \textit{``specialists''}, particularly designed/fine-tuned for tool-use benchmarks in the same papers that proposed the models.
We evaluate them on their specialized benchmarks.
Additionally, we evaluate Llama and Mistral-7B, two \textit{``generalist''} instruction-tuned models to demonstrate that \method can also work on base models that are not particularly designed for tool use.
These two models are not fine-tuned on the benchmark's training sets.
Therefore, their initial performance without the \method is very poor compared to their specialist counterparts.

\textbf{ToolLLM \citep{qin2023toolllm}.}
ToolLLM is a specialist LLaMA-7B model fine-tuned to use tools on RapidAPI (\url{https://rapidapi.com/}).
Given each task, ToolLLM is prompted with the documentation of relevant tools.
It is then instructed to generate a natural language rationale, a tool to use, and the tool's inputs.
This process continues for several iterations until the path of tool calls leads to an answer or the model gives up.
ToolLLM can either generate a single path of tool calls (denoted as \textit{ToolLLM+ReAct}) or run a tree search to find the best path (denoted as \textit{ToolLLM+DFSDT}).
Note that ToolLLM is fine-tuned with the \textit{same} formatted data synthesized with the \textit{same} process as its evaluation benchmark, ToolEval.

\textbf{ToolkenGPT \citep{hao2023toolkengpt}.}
ToolkenGPT is a specialist Llama-33B model with an additional vocabulary of tokens learned for tools.
The original weights of Llama are frozen while the additional tokens are learned.
These tokens for tools, or \textit{toolkens}, each correspond to a tool.
The text spans of tool use in the training data are replaced with toolkens, so that the model can learn when to start a tool call.
Although ToolkenGPT is much more efficient than full fine-tuning a model, it still needs to learn the representations of all tools in the evaluation set.

\textbf{RestGPT \citep{song2023restgpt}.}
RestGPT learns to use RESTful APIs from in-context tool documentation.
RestGPT utilizes several different LLM-based modules to make natural language plans, select which APIs to use, generate tool calls and parse results from them.

\textbf{Mistral-Instruct \citep{jiang2023mistral} and LLaMA \citep{touvron2023llama}.}
Mistral-Instruct-7B and LLaMA-7B are generalist 7B LLMs tuned without tool-specific fine-tuning.
When directly evaluated on ToolEval with syntax constraints in the prompt, Mistral cannot solve any tests.
This suggests that Mistral is not able to follow syntax constraints when prompted.
The same is true for LLaMA-7B.
We choose them as base models to show when syntax constraints are taken care of by \method; even a generalist model can perform well on tool-using.

\subsection{Benchmarks and Metrics}
\label{sec:bench}

We evaluate the baselines and \method on four benchmarks -- ToolEval \citep{qin2023toolllm}, FuncQA \citep{hao2023toolkengpt}, KAMEL \citep{kalo2022kamel}, and RestBench \citep{song2023restgpt}.
We evaluate the performance of all base models with and without \method.
Other than correctness metrics, we also measure \textbf{syntax error rate}, the proportion of tasks on which the models make syntax errors.
Our experiments are conducted on NVIDIA A6000 GPUs within 60 GPU hours.

\textbf{ToolEval \citep{qin2023toolllm}.} ToolEval is a dataset proposed in the ToolLLM paper. Tasks in it involve 10000+ publicly available REST APIs.
We use its more complex subsets to evaluate our method -- I2-Category and I3-Instruction.
They contain tasks that need complex and unseen tools from multiple categories to solve.
On average, a task in these subsets needs more than 7 tools to solve.
ToolEval has two main metrics: \textbf{pass rate} measures the percentage of tasks for which the model reaches an answer within limited reasoning steps. \textbf{win rate}  compares the quality and correctness of the models' answers to the reference answers from ChatGPT. \citet{qin2023toolllm} finds that these automatic metrics have a high correlation of 75.8\% with human annotators.
Since ToolEval requires extensive prompting of tool documentation, we measure \textbf{tok / tool}, the average number of tokens required for each tool, to see how much \method can shorten the prompts.

\textbf{FuncQA \citep{hao2023toolkengpt}.} FuncQA tests LLMs' ability in numerical reasoning tasks with 68 math problems.
LLMs are required to produce a numerical answer using a few of the 13 arithmetic operations as tools (e.g. \texttt{multiply}, \texttt{power}, \texttt{lcm}).
The accuracy is determined by measuring the percentage of problems for which a correct answer is produced, with a 0.1\% error tolerance.
On average, a problem in FuncQA requires 2.78 tool calls to solve.
Following \cite{hao2023toolkengpt}, we report results of other baselines, including ChatGPT without tools, LLaMA with chain-of-thought and tools, LLaMA with ReAct and tools.

\textbf{KAMEL \citep{kalo2022kamel}.} KAMEL is a question-answering dataset containing a total of 234 knowledge relations that resemble the characteristics of APIs (e.g. \texttt{number\_of\_children}).
The tools in KAMEL are also more complex and diverse because their number of arguments varies from 1 to 3, and their types include strings, locations, dates, numbers, and other ad-hoc types.

\textbf{RestBench \citep{song2023restgpt}.}
RestBench consists of tasks in real-world scenarios, including TMDB, a movie website, and Spotify, an online music player.
These tasks directly come from real-user instructions and require multiple tools in the form of RESTful APIs to solve.
We use the correct path rate (CP\%) proposed by the original paper as the metric to measure accuracy.
Correct path rate is the proportion of outputs that contain the correct tool call path annotated by humans.

\begin{table*}[t!]
\small
\centering
\caption{When applied to various baselines on different benchmarks, \method significantly improves the model generations with fewer tokens in the prompt.
It also completely eliminates all syntax errors.
``err.'' is short for syntax error rate.
``tok'' is short for the average number of tokens for each tool.
}
\label{tab:main}
\begin{tabular}{rcccccccc}
\toprule
                & \multicolumn{4}{c}{\textbf{ToolEval : I2-Category}}                      & \multicolumn{4}{c}{\textbf{ToolEval: I3-Instruction}}                   \\ 
                & win\% $\uparrow$ & pass\% $\uparrow$  & tok $\downarrow$& err.\% $\downarrow$ & win\% $\uparrow$ & pass\% $\uparrow$  & tok $\downarrow$& err.\% $\downarrow$ \\ \midrule
ToolLLM + ReAct& 36.5 & 30.5 & 827 & 21 & 49 & 22 & 1380 & 32 \\
\rowcolor{myhighlight}
+ \method       & 46.5  & 47.5  & 397& 0 & 61  &  48  & 627 &  0                  \\
ToolLLM + DFSDT& 40.5 & 64.5 & 827  & 44    & 41 & 58 & 1380 & 49                \\
\rowcolor{myhighlight}
+ \method & 50.5  & 69 & 
397& 0 &  49 & 59 & 627&  0                 \\
Mistral + ReAct& \textbf{\textcolor{red}{0}} & \textbf{\textcolor{red}{0}}& 827 & 92.5 & \textbf{\textcolor{red}{0}}& \textbf{\textcolor{red}{0}}& 1380 & 93
\\
\rowcolor{myhighlight}
+ \method & 41 & 26 & 397& 0& 53 & 32 & 627& 0\\
Mistral + DFSDT& \textbf{\textcolor{red}{0}}& \textbf{\textcolor{red}{0}}& 827 & 100 & \textbf{\textcolor{red}{0}}& \textbf{\textcolor{red}{0}}& 1380 & 100
\\
\rowcolor{myhighlight}
+ \method  & 52  & 50.5  & 397& 0& 60 & 35 & 627& 0\\ \midrule
ChatGPT + ReAct & n.a. & 39 & 827 & 5  & n.a. & 23 & 1380 & 9
\\
ChatGPT + DFSDT & 63.0 & 64.5 & 827 & 28 & 70 & 60 & 1380 & 47
\\
GPT-4 + ReAct & 53.5 & 67.5 & 827 & 4 & 71 & 40 & 1380  & 2 
\\
GPT-4 + DFSDT & 57 & 69.5 & 827 & 4  & 73 & 59 & 1380 & 5
\\ \toprule
& \multicolumn{4}{c}{\textbf{FuncQA}}                      & \multicolumn{4}{c}{\textbf{KAMEL}}
\\
& \multicolumn{2}{c}{accuracy \% $\uparrow$} & \multicolumn{2}{c}{err.\%  $\downarrow$} & \multicolumn{2}{c}{accuracy \% $\uparrow$} &  \multicolumn{2}{c}{err.\%  $\downarrow$} \\ \midrule
LLaMA& \multicolumn{2}{c}{6} & \multicolumn{2}{c}{27.9} & \multicolumn{2}{c}{8.2} & \multicolumn{2}{c}{n.a.}\\
\rowcolor{myhighlight}
+ \method & \multicolumn{2}{c}{13.2 }  & \multicolumn{2}{c}{0}& \multicolumn{2}{c}{42.4 } &\multicolumn{2}{c}{n.a.}    \\ 
ToolkenGPT & \multicolumn{2}{c}{10.3} & \multicolumn{2}{c}{27.9}& \multicolumn{2}{c}{25.4} & \multicolumn{2}{c}{n.a.}   \\
\rowcolor{myhighlight}
+ \method & \multicolumn{2}{c}{13.2 } & \multicolumn{2}{c}{0  } & \multicolumn{2}{c}{n.a.} & \multicolumn{2}{c}{n.a.}   \\
\toprule
& \multicolumn{4}{c}{\textbf{RestBench: Spotify}}                      & \multicolumn{4}{c}{\textbf{RestBench: TMDB}}
\\
& \multicolumn{2}{c}{correct path \% $\uparrow$} & \multicolumn{2}{c}{err.\% $\downarrow$} & \multicolumn{2}{c}{correct path \% $\uparrow$}  & \multicolumn{2}{c}{err.\% $\downarrow$}
\\ \midrule 
Vicuna + RestGPT & \multicolumn{2}{c}{20.6}  & \multicolumn{2}{c}{7.4} & \multicolumn{2}{c}{15} & \multicolumn{2}{c}{23} \\
\rowcolor{myhighlight}
+\method & \multicolumn{2}{c}{36} &  \multicolumn{2}{c}{0 } & \multicolumn{2}{c}{23} & \multicolumn{2}{c}{0 }  \\ \midrule
ChatGPT & \multicolumn{2}{c}{72.3} & \multicolumn{2}{c}{3.6} & \multicolumn{2}{c}{65} &\multicolumn{2}{c}{3} \\ \bottomrule
\end{tabular}
\end{table*}

\subsection{Results}
\label{sec:main}

We list our main results in \autoref{tab:main}.
Each green row contains the performance of \method applied to the base model in the previous row. Note that on KAMEL, syntax error rate is not available. This is because KAMEL tests the model's ability to select the right tool and does not involve tool arguments and execution. For the same reason, ToolkenGPT + ToolDec is also not available.
Also, the win rates on ToolEval are not available for ChatGPT + ReAct, because it is the baseline compared against when computing the win rates.
Several observations can be made about the results:

\textbf{\method leads to significant improvements of all base models across multiple benchmarks. }
On ToolEval, ToolDec improves win rates and pass rates substantially on ToolLLM with either the ReAct or DFSDT strategies. The win rate is improved by 10 points, while the pass rate is improved by 12 on average.
Particularly noteworthy is the performance of ToolLLM+DFSDT 
 with \method, which not only outperforms ChatGPT but also achieves performance on par with GPT-4.
Similarly significant improvements are observed across other benchmarks on other base models as well.

\textbf{\method helps generalist models to match or outperform similar-sized specialist models.}
Without \method, Mistral-7B cannot pass a single task on ToolEval, while ToolLLM has an average >40\% win rate and >40\% pass rate.
With \method, Mistral-7B's performance on ToolEval is on par with ToolLLM + \method and even better in several cases and metrics.
The same is true for FuncQA and KAMEL.
When enhanced with \method, the generalist, pre-trained LLaMA-7B model's performance gets 2.2x better on FuncQA and 5.1x better on KAMEL, matching or beating the specialist ToolkenGPT.

\textbf{\method gets much better performance with shorter prompts.}
On ToolEval, the average number of tokens required for each tool is reduced by 58\% and 69\% for I2-Category and I3-Instruction.
With shorter prompts, \method is able to include more tools in the inventory without exceeding the context limit of LLMs.

\textbf{All models have syntax errors. Fine-tuning or prompting can't eliminate them, but \method can.}
When evaluating on the ToolEval: I3-Instruction dataset, ToolLLM + DFSDT—despite being fine-tuned with data formatted identically to that of the test data—still produces syntax errors in nearly half (49\%) of the instances. Mistral is even worse, with syntax errors occurring in over 90\% of its responses. This high error rate probably explains its directly impacted its accuracy being exactly 0. Even advanced models like ChatGPT and GPT-4, known for their capabilities, were not immune to syntax mistakes during tool interactions.
The same is true for other benchmarks and other base models.
When \method is applied, both specialist models like ToolLLM and generalist models stop making syntax errors.

\subsection{Ablation Study}
\label{sec:abl}

\textbf{\method v.s. Tool-Specialized Fine-Tuning.} 
We argue that it's more ideal to use \method on generalist LLMs instead of fine-tuning specialist LLMs to use tools.
To demonstrate that, we evaluate both the specialist ToolLLM and the generalist Llama-2-chat and Mistral-Ins on other reasoning and coding benchmarks, including GSM8K \citep{cobbe2021gsm8k}, HumanEval \citep{chen2021evaluating}, MBPP \citep{austin2021program}, and BigBenchHard \citep{suzgun2022challenging}.
The details of these evaluations can be found in Appendix \ref{app:eval}.
As reported in \autoref{tab:abl}, although the specialist ToolLLM can have comparable performance on tool-related benchmarks as \method on generalist LLMs, its performance on general reasoning or coding benchmarks is unreasonably bad.
On GSM8K, it's solving only 1.2\% of the problems, while Llama-2-chat, the generalist model with the same pertaining, solves 23.1\%.
This suggests that tool-specialized fine-tuning can seriously degrade the model's general ability, probably due to catastrophic forgetting.

\begin{table*}[tb!]
\small
\centering
\caption{
While tool-specialized models, such as ToolLLM, can perform as well as generalist models + \method ($\dagger$), their performance on other reasoning benchmarks is much lower (underlined).
}
\label{tab:abl}
\begin{tabular}{rccccccc}
\toprule
& ToolEval (I2-Cat) & ToolEval (I2-Ins) & GSM8K & HumanEval & MBPP & BBH \\
\midrule
ToolLLM-7B & 36.5& 49.0 & \textbf{\color{red}{\underline{1.2}}} & \textbf{\color{red}{\underline{2.4}}} & \textbf{\color{red}{\underline{5.3}}} & \textbf{\color{red}{\underline{28.1}}} \\
Llama-2-chat-7B $\dagger$ & 39.0 & 51.0 & 23.1 & 14.6 & 29.6 & 39.9 \\
Mistral-Ins-7B $\dagger$ & 41.0 & 53.0 & 42.0 & 43.9 & 43.4 & 50.6 \\
\bottomrule
\end{tabular}
\end{table*}

\begin{figure}[t!]
    \centering
    \begin{subfigure}[b]{0.45\textwidth}
        \includegraphics[width=\textwidth]{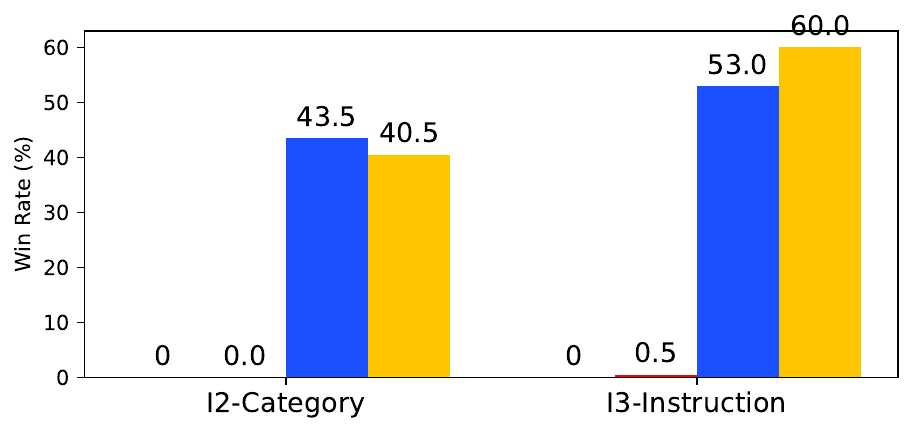}
        \caption{Win rates of Mistral on I2-Category and I3-Instruction under four different settings.}
        \label{fig:mistral_abl}
    \end{subfigure}
    \vspace{3pt} %
    \begin{subfigure}[b]{0.45\textwidth}
        \includegraphics[width=\textwidth]{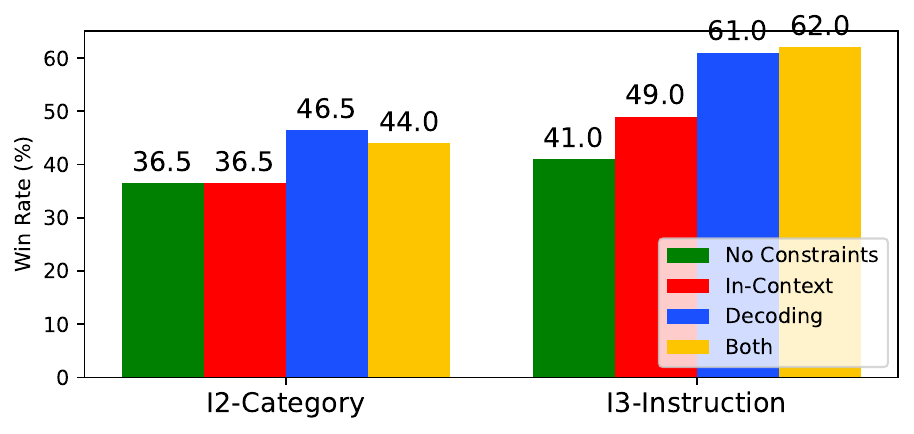}
        \caption{Win rates of ToolLLM on I2-Category and I3-Instruction under four different settings.}
        \label{fig:tolllm_abl}
    \end{subfigure}
    \caption{Win rates of Mistral and ToolLLM under different settings of syntax constraints. Syntax constraints are much more helpful as decoding constraints than in-context descriptions. %
    } \label{fig:abl_cons}
    \vspace{-10pt}
\end{figure}

\textbf{\method vs. Prompting Syntax Constraints}.
We argue that removing syntax constraints in prompts doesn't harm \method's performance.
We conduct a fine-grained ablation study on two approaches (prompting and decoding) to incorporate syntax constraints with two base models -- ToolLLM and Mistral. We consider four different settings of syntax constraints: ``no prompting, no decoding'', ``only decoding'', ``only prompting'', ``prompting and decoding''.
We evaluate the win rate of ToolLLM and Mistral under these four settings and report the results in \autoref{fig:abl_cons}.
The following observations can be made: 1) Decoding time syntax constraints are much more helpful than in-prompt syntax constraints.
For both models, the improvements from constrained decoding are much larger than those from prompting syntax constraints.
2) Using decoding time constraints is mostly enough. Extra in-context constraints offer little help.
The gaps between the third setting, decoding constraints, and the fourth setting, both constraints are very small.

%% file: sections/070_conclusion.tex
We propose \method, a constrained decoding algorithm to guide LLMs to use external tools.
It guarantees that LLMs do not make syntax errors when generating a tool call.
\method is complementary to existing approaches and improves the performance of all the base LLMs we tested.
Surprisingly, \method can enable generalist LLMs to be as good as or even better than tool-specialized models.
Since very capable language models, such as GPT-4, can still make syntax errors, we believe that \method can be used in some security-sensitive areas to avoid syntax errors, thus having a positive societal impact.
However, we do acknowledge that syntax error-free doesn't mean error-free.
LLMs with \method can still make mistakes beyond syntax and should be used with care.

%% file: sections/100_app.tex
\section{Appendix}

\subsection{Examples of Tool Syntax Errors}

\label{app:example}
\begin{figure*}[h]
  \centering
  \includegraphics[scale=0.75]{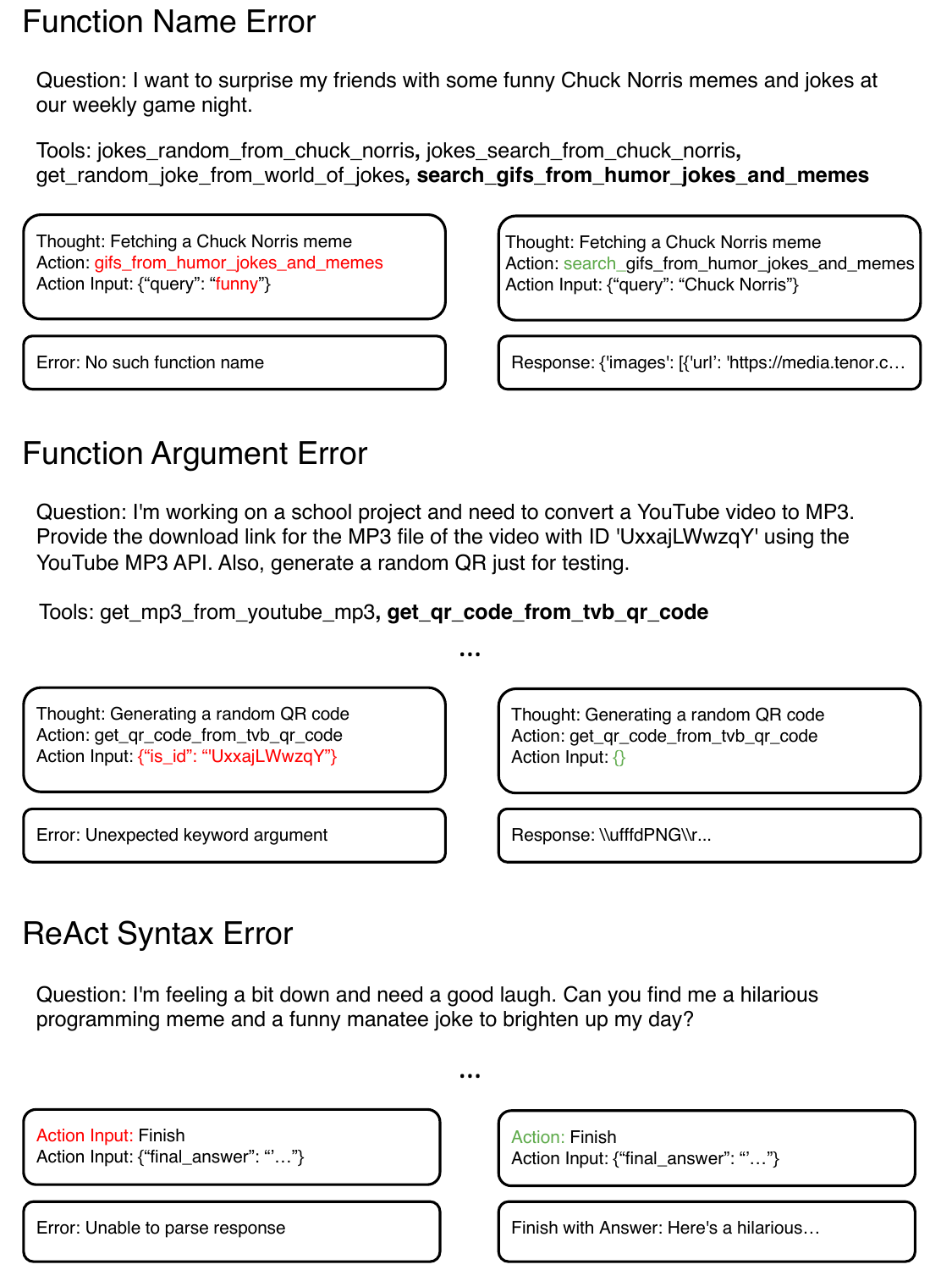}
  \caption{\label{fig:elimination_example}\method can prevent function name error, function argument error, and invalid ReAct syntax on ToolLLM.}
\end{figure*}

\newpage
\subsection{Algorithm Pseudocode}
\begin{algorithm}
\caption{Finite-State Machine Guided Decoding for Language Models}
\begin{algorithmic}
\STATE {\bfseries Input:} data $x_i$, size $m$ \\
A DFSM defined by $(S,V,g,s_0,R)$;\\
A language model $M$ that produces the distribution of the next token given a prefix string;\\
An initial string of tokens $x_{1..k}$, which represents the prompt from the user.
\STATE {\bfseries Output:} data $x_i$, size $m$
\STATE $s\leftarrow s_0$
\WHILE {$s\not \in F$}
    \STATE $V_{s}\leftarrow \{a | a \in V \land g(s, a)\text{ is defined} \}$
    \STATE $P(x_{k+1}|x_{1..k}) \leftarrow M(x_{1..k})$
    \STATE $\tilde P(x=a|x_{1..k},s)\leftarrow$
    \STATE $\left\{
    \begin{aligned}&\tfrac{P(x=a|x_{1..k})}{\sum_{a'\in V_s}P(x=a'|x_{1..k})},\ &a\in V_s\\ &0, & \text{otherwise}\end{aligned}\right.$
    \STATE $x\sim \tilde P(x|x_{1..k},s)$
    \STATE $x_{k+1} \leftarrow x$
    \STATE $k\leftarrow k+1$
    \STATE $s\leftarrow g (s, x)$
\ENDWHILE
\STATE {\bf return} $x_{1..k}$
\end{algorithmic}\label{algo}
\end{algorithm}

\subsection{Prompt to Remove Syntax Constraints}
\label{app:prompt}
\begin{verbatim}
The user will give you a list of functions in JSON and you will simplify their 
descriptions. For each function, write a concise but semantically rich description 
of its purpose but you do not need to mention the tool it belongs to. List out the 
parameters, one per line, and write a concise but semantically rich description.
You do not need to include syntactical  information (such as parameter type) but 
please include the example value if available.  Your response must be in plain text 
and wrapped in a code block.
            
Here is an example. Please follow the same syntax:

1. airport_arrivals_for_flight_fare_search

   Description: Retrieves information about arriving flights.
   Parameters:
   - airportcode: Airport code (Example: LHR).
   - carriercode: Airline carrier code (Optional).
   - date: Date for checking arrivals (Optional).
\end{verbatim}

\subsection{Evaluation Details}
\label{app:eval}

We report win rates under the ReAct setting on ToolEval.
We use lm-eval-harness (\url{https://github.com/EleutherAI/lm-evaluation-harness}) to evaluate LLMs on GSM8K (5-shot) and BigBenchHard (3-shot, CoT) using the default parameters.
We use eval-plus (\url{https://github.com/evalplus/evalplus/}) to evaluate LLMs on HumanEval and MBPP with greedy decoding and report pass@1 results.